\documentclass[11pt,a4paper]{article}
\usepackage[hyperref]{acl2021}
\usepackage{times}
\usepackage{latexsym}

\usepackage{microtype}
\usepackage[utf8]{inputenc} % allow utf-8 input
\usepackage[T1]{fontenc}    % use 8-bit T1 fonts
\usepackage{url}            % simple URL typesetting
\usepackage{booktabs}       % professional-quality tables
\usepackage{amsfonts}       % blackboard math symbols
\usepackage{nicefrac}       % compact symbols for 1/2, etc.
\usepackage{amsmath}

\usepackage{graphicx}
\usepackage{bm, upgreek}
\usepackage{amssymb}
\usepackage{array, booktabs}
\usepackage{multirow}
\usepackage{mathtools}
\usepackage{arydshln}
\usepackage{tabularx}
\usepackage{color}
\usepackage{soul}
\usepackage{caption}
\usepackage{xcolor}
\usepackage{pifont}% http://ctan.org/pkg/pifont
\usepackage{comment}
\usepackage{cleveref}
\usepackage{enumitem}
\usepackage[ruled,vlined]{algorithm2e}

\crefformat{section}{\S#2#1#3} % see manual of cleveref, section 8.2.1
\crefformat{subsection}{\S#2#1#3}
\crefformat{subsubsection}{\S#2#1#3}

\newcommand{\cmark}{\ding{51}}%
\newcommand{\xmark}{\ding{55}}%

\definecolor{brickred}{rgb}{0.8, 0.25, 0.33}
\definecolor{blue-green}{rgb}{0.0, 0.87, 0.87}
\definecolor{celestialblue}{rgb}{0.29, 0.59, 0.82}
\definecolor{cerulean}{rgb}{0.0, 0.48, 0.65}
\definecolor{ceruleanblue}{rgb}{0.16, 0.32, 0.75}

\newcommand{\ours}{DensePhrases}

\newcommand{\documentset}{\mathcal{D}}
\newcommand{\phraseset}{\mathcal{S}(\mathcal{D})}
\newcommand{\phraseinp}{\mathcal{S}(p)}
\newcommand{\wordset}{\mathcal{W}(\mathcal{D})}
\newcommand{\phrasedump}{\mf{H}}
\newcommand{\traincorpus}{\mathcal{C}}
\newcommand{\lm}{\mathcal{M}}

\newcommand\ti[1]{\textit{#1}}

\newcommand\tf[1]{\textbf{#1}}
\newcommand\ttt[1]{\texttt{#1}}
\newcommand\mf[1]{\mathbf{#1}}

\DeclareMathOperator*{\argmax}{argmax}

\newcommand{\gpu}[1]{\textcolor{brickred}{#1}}
\newcommand{\cpu}[1]{\textcolor{ceruleanblue}{#1}}

\newcommand{\draftonly}[1]{#1}
\newcommand{\draftcomment}[3]{\draftonly{\textcolor{#2}{{{[#1: #3]}}}}}

\newcommand{\jinhyuk}[1]{\draftcomment{Jinhyuk}{blue}{#1}}

\aclfinalcopy % Uncomment this line for the final submission
\def\aclpaperid{1323} %  Enter the acl Paper ID here

\title{Learning Dense Representations of Phrases at Scale}
\author{
  Jinhyuk Lee$^{1,2}$\Thanks{ Work partly done while visiting Princeton University.}\quad Mujeen Sung$^{1}$ \quad Jaewoo Kang$^{1}$ \quad Danqi Chen$^{2}$\\
  Korea University$^{1}$\quad Princeton University$^{2}$\\
  \texttt{\{jinhyuk\_lee,mujeensung,kangj\}@korea.ac.kr} \\
  \texttt{danqic@cs.princeton.edu} \\}

\date{}

\begin{document}
\maketitle

%!TEX root = ../acl2021.tex

\begin{abstract}
Open-domain question answering can be reformulated as a phrase retrieval problem, without the need for processing documents on-demand during inference~\citep{seo2019real}.
However, current phrase retrieval models heavily depend on sparse representations and still underperform retriever-reader approaches.
In this work, we show for the first time that we can learn \ti{dense representations of phrases} alone that achieve much stronger performance in open-domain QA.
% Our approach includes (1) learning query-agnostic phrase representations via question generation and distillation; (2) novel negative-sampling methods for global normalization; (3) query-side fine-tuning for transfer learning.
We present an effective method to learn phrase representations from the supervision of reading comprehension tasks, coupled with novel negative sampling methods.
We also propose a query-side fine-tuning strategy, which can support transfer learning and reduce the discrepancy between training and inference.
% Our approach includes (1) learning effective phrase representations from reading comprehension datasets, coupled with novel negative sampling methods; (2) query-side fine-tuning for reducing the discrepancy between training and testing as well as supporting trasnfer learning.
On five popular open-domain QA datasets, our model \ti{DensePhrases} improves over previous phrase retrieval models by $15\%$--$25\%$ absolute accuracy and matches the performance of state-of-the-art retriever-reader models.
Our model is easy to parallelize due to pure dense representations and processes more than 10 questions per second on CPUs.
Finally, we directly use our pre-indexed dense phrase representations for two slot filling tasks, showing the promise of utilizing {DensePhrases} as a dense knowledge base for downstream tasks.\footnote{Our code is available at \url{https://github.com/princeton-nlp/DensePhrases}.}

\end{abstract}

%!TEX root = ../acl2021.tex

\section{Introduction}
\label{sec:intro}

Open-domain question answering (QA) aims to provide answers to natural-language questions using a large text corpus~\citep{voorhees1999trec,ferrucci2010building,chen2020open}.
While a dominating approach is a two-stage retriever-reader approach~\citep{chen2017reading,lee2019latent,guu2020realm,karpukhin2020dense},
we focus on a recent new paradigm solely based on \textit{phrase retrieval}~\citep{seo2019real,lee2020contextualized}.
Phrase retrieval highlights the use of phrase representations and finds answers purely based on the similarity search in the vector space of phrases.\footnote{Following previous work~\citep{seo2018phrase}, `phrase' denotes any contiguous segment of text up to $L$ words (including single words), which is not necessarily a linguistic phrase.}
Without relying on an expensive reader model for processing text passages, it has demonstrated great runtime efficiency at inference time.
% Table~\ref{tab:category} compares the two approaches in detail.

Despite great promise, it remains a formidable challenge to build vector representations for every single phrase in a large corpus.
Since phrase representations are decomposed from question representations, they are inherently less expressive than cross-attention models~\cite{devlin2019bert}.
% , even in a single-passage setting.
% they are often less expressive than cross-attention models~\cite{devlin2019bert}.
% ---this challenge brings the \textit{decomposability gap} as stated in~\citep{seo2018phrase,seo2019real}.
Moreover, the approach requires retrieving answers correctly out of {billions} of phrases (e.g., $6 \times 10^{10}$ phrases in English Wikipedia), making the scale of the learning problem difficult.
% which are more than four orders of magnitude larger than the number of documents in Wikipedia.
Consequently, existing approaches heavily rely on sparse representations for locating relevant documents and paragraphs while still falling behind retriever-reader models~\citep{seo2019real,lee2020contextualized}.

%!TEX root = ../acl2021.tex

\begin{table*}[t]
\label{table:open_qa_results}
\begin{center}
\centering
\resizebox{2.0\columnwidth}{!}{%
\begin{tabular}{llccccc}
\toprule
\multirow{2}{*}{\textbf{Category}} & \multirow{2}{*}{\textbf{Model}} & \multirow{2}{*}{\textbf{Sparse?}} & {\textbf{Storage}} & {\textbf{\#Q/sec}} & \textbf{NQ}  & \textbf{SQuAD} \\
& & & \multicolumn{1}{c}{(GB)} & (\gpu{GPU}, \cpu{CPU}) & (Acc) & (Acc) \\
\midrule
\multirow{5}{*}{Retriever-Reader} & DrQA~\citep{chen2017reading} & \cmark & 26 & \gpu{1.8}, \cpu{0.6} & - & 29.8 \\
% & BM25 + BERT ~\citep{lee2019latent} & \cmark & - & - & 26.5 & 33.2 \\
& BERTSerini~\citep{yang2019end} & \cmark & 21 & \gpu{2.0}, \cpu{0.4} & - & \tf{38.6} \\
& ORQA~\citep{lee2019latent} & \xmark & 18 & \gpu{8.6}, \cpu{1.2} & 33.3 & 20.2 \\
& REALM$_\text{News}$~\citep{guu2020realm} & \xmark & 18 & \gpu{8.4}, \cpu{1.2} & 40.4 & -\\
& DPR-multi~\citep{karpukhin2020dense}& \xmark & 76 & \gpu{0.9}, \cpu{0.04} & \tf{41.5} & 24.1 \\
\midrule
% Reader Only &  T5-11B + SSM~\citep{roberts2020much} & \xmark & 42 & - & 34.8 & - \\
% \midrule
\multirow{3}{*}{Phrase Retrieval} & DenSPI~\citep{seo2019real} & \cmark & 1,200 & \gpu{2.9}, \cpu{2.4} & 8.1 & 36.2\\
& DenSPI + Sparc~\citep{lee2020contextualized} & \cmark & 1,547 & \gpu{2.1}, \cpu{1.7} & 14.5 & \tf{40.7} \\
& \textbf{\ours~(Ours)} & \xmark & 320 & \gpu{20.6},	\cpu{13.6} & \tf{40.9} & 38.0 \\
\bottomrule
\end{tabular}
}
\end{center}\vspace{-0.1cm}
\caption{Retriever-reader and phrase retrieval approaches for open-domain QA. The \textit{retriever-reader} approach retrieves a small number of relevant documents or passages from which the answers are extracted.
The \textit{phrase retrieval} approach retrieves an answer out of billions of phrase representations pre-indexed from the entire corpus. \Cref{apdx:server} provides detailed benchmark specification. The accuracy is measured on the test sets in the open-domain setting.  NQ: Natural Questions.
}\vspace{-0.4cm}
\label{tab:category}
\end{table*}

In this work, we investigate whether we can build fully dense phrase representations at scale for open-domain QA.
First, we aim to learn strong phrase representations from the supervision of reading comprehension tasks.
We propose to use data augmentation and knowledge distillation to learn better phrase representations within a single passage.
We then adopt negative sampling strategies such as in-batch negatives~\citep{henderson2017efficient,karpukhin2020dense}, to better discriminate the phrases at a larger scale.
% First, we attribute the cause of the decomposability gap to the sparsity of training data.
% We close this gap by generating questions for every answer phrase, as well as distilling knowledge from query-dependent models (\Cref{sec:piqa}).
% {Second}, we use negative sampling strategies such as in-batch negatives~\citep{henderson2017efficient,karpukhin2020dense}, to approximate global normalization.
Here, we present a novel method called \ti{pre-batch negatives}, which leverages preceding mini-batches as negative examples to compensate the need of large-batch training.
Lastly, we present a {query-side fine-tuning strategy} that drastically improves phrase retrieval performance and allows for transfer learning to new domains, without re-building billions of phrase representations.

As a result, all these improvements lead to a much stronger phrase retrieval model, without the use of \ti{any} sparse representations (Table~\ref{tab:category}). We evaluate our model, \ti{DensePhrases}, on five standard open-domain QA datasets and achieve much better accuracies than previous phrase retrieval models~\citep{seo2019real,lee2020contextualized}, with 15\%--25\% absolute improvement on most datasets.
Our model also matches the performance of state-of-the-art retriever-reader models~\citep{guu2020realm,karpukhin2020dense}.
Due to the removal of sparse representations and careful design choices, we further reduce the storage footprint for the full English Wikipedia from 1.5TB to 320GB, as well as drastically improve the throughput.
% The removal of sparse representations and careful design choices reduce the storage footprint of phrase retrieval from 1.5TB to 320GB, as well as drastically improving the throughput.

Finally, we envision that DensePhrases acts as a neural interface for retrieving phrase-level knowledge from a large text corpus.
% As such, it can be integrated into other knowledge-intensive NLP tasks beyond question answering.
To showcase this possibility, we demonstrate that we can directly use DensePhrases for fact extraction, without re-building the phrase storage.
With only fine-tuning the question encoder on a small number of subject-relation-object triples, we achieve state-of-the-art performance on two slot filling tasks~\citep{petroni2020kilt}, using less than 5\% of the training data.
% \jinhyuk{5K is 0.2\% for T-REx and 3.4\% for ZsRE }

% on a small number of subject-relation-object triples on the query representations alone,
% \footnotetext[2]{Since we benchmark each model with a batch size of $64$ as detailed in~\Cref{apdx:server}, the speedup in previous phrase retrieval models~\cite{seo2019real,lee2020contextualized} is smaller than previously claimed, although they still performs much faster on CPUs.}

%!TEX root = ../acl2021.tex

\section{Background}
\label{sec:background}

% \subsection{Open-domain QA}

We first formulate the task of open-domain question answering for a set of $K$ documents $\documentset=\{d_1, \dots, d_K\}$.
We follow the recent work~\cite{chen2017reading,lee2019latent} and treat all of English Wikipedia as $\documentset$, hence $K \approx 5 \times 10^6$.
However, most approaches---including ours---are generic and could be applied to other collections of documents.

The task aims to provide an answer $\hat{a}$ for the input question $q$ based on $\documentset$.
In this work, we focus on the extractive QA setting, where each answer is a segment of text, or a \ti{phrase}, that can be found in $\documentset$.
Denote the set of phrases in $\documentset$ as $\phraseset$ and each phrase $s_{k} \in \phraseset$ consists of contiguous words $w_{\ttt{start}(k)}, \ldots, w_{\ttt{end}(k)}$ in its document $d_{\ttt{doc}(k)}$.
In practice, we consider all the phrases up to $L = 20$ words in $\documentset$ and $\phraseset$ comprises a large number of $6 \times 10^{10}$ phrases.
An extractive QA system returns a phrase $\hat{s} = \argmax_{s \in \mathcal{S}(\documentset)} f(s | \documentset, q)$ where $f$ is a scoring function. The system finally maps $\hat{s}$ to an answer string $\hat{a}$: $\ttt{TEXT}{(\hat{s})}=\hat{a}$ and the evaluation is typically done by comparing the predicted answer $\hat{a}$ with a gold answer $a^*$.

%\jinhyuk{moved to here}
Although we focus on the extractive QA setting, recent works propose to use a generative model as the reader~\citep{lewis2020retrieval,izacard2020leveraging}, or learn a closed-book QA model~\citep{roberts2020much}, which directly predicts answers  without using an external knowledge source. The extractive setting provides two advantages: first, the model directly locates the source of the answer, which is more interpretable, and second, phrase-level knowledge retrieval can be uniquely adapted to other NLP tasks as we show in~\S\ref{sec:slot_filling}.

\paragraph{Retriever-reader.}
A dominating paradigm in open-domain QA is the retriever-reader approach~\cite{chen2017reading,lee2019latent,karpukhin2020dense}, which leverages a first-stage document retriever $f_{\text{retr}}$ and only reads top $K' \ll K$ documents with a reader model $f_{\text{read}}$. The scoring function $f(s \mid \mathcal{D}, q) $ is decomposed as:
\vspace{-0.3em}
\begin{equation}
\begin{split}
     f(s \mid \mathcal{D}, q) & =  f_{\text{retr}}(\{d_{j_1}, \ldots, d_{j_{K'}}\} \mid \mathcal{D}, q) \\ & \times f_{\text{read}}(s \mid \{d_{j_1}, \ldots, d_{j_{K'}}\}, q),
\end{split}
\end{equation}
where $\{j_1, \ldots, j_{K'}\} \subset \{1, \ldots, K\}$ and if $s \notin \mathcal{S}(\{d_{j_1}, \ldots, d_{j_{K'}}\})$, the score will be 0.
It can easily adapt to passages and sentences~\cite{yang2019end,wang2019multi}.
However, this approach suffers from error propagation when incorrect documents are retrieved and can be slow as it usually requires running an expensive reader model on every retrieved document or passage at inference time.

\begin{comment}
\paragraph{Reader Only}
A second paradigm, the reader only approach~\citep{roberts2020much} (also known as a closed-book models\jinhyuk{add EaE?} merely depends on parameters of language models ($f_{\text{LM}}$) to store all the factual knowledge:
\begin{equation}
\begin{split}
     f(a \mid \mathcal{D}, q) = f_{\text{LM}}(a \mid q).
\end{split}
\end{equation}
This approach requires gigantic language models (e.g., T5 models with 11G parameters) to achieve competitive performance. It is also known that these models perform more question memorization than generalization~\cite{lewis2020question}.
\end{comment}

\paragraph{Phrase retrieval.}
\citet{seo2019real} introduce the phrase retrieval approach that encodes phrase and question representations \textit{independently} and performs similarity search over the phrase representations to find an answer.
Their scoring function $f$ is computed as follows:
\vspace{-0.5em}
\begin{equation}\label{eqn:decomp}
\begin{split}
    f(s \mid \documentset, q) = {E}_{s}(s, \documentset)^\top {E}_{q}(q),
\end{split}
\end{equation}
where $E_s$ and $E_q$ denote the phrase encoder and the question encoder respectively.
As $E_s(\cdot)$ and $E_q(\cdot)$ representations are decomposable, it can support maximum inner product search (MIPS) and improve the efficiency of open-domain QA models.
Previous approaches~\cite{seo2019real,lee2020contextualized} leverage both dense and sparse vectors for phrase and question representations by taking their concatenation: $E_{s}(s, \documentset) = [E_{\text{sparse}}(s, \documentset),  E_{\text{dense}}(s, \documentset)].$\footnote{\newcite{seo2019real} use sparse representations of both paragraphs and documents  and \newcite{lee2020contextualized} use contextualized sparse representations conditioned on the phrase.}
However, since the sparse vectors are difficult to parallelize with dense vectors, their method  essentially conducts sparse and dense vector search separately.
% is not purely implemented as MIPS in practice while requiring more storage.
The goal of this work is to only use dense representations, i.e., $E_{s}(s, \mathcal{D}) = E_{\text{dense}}(s, \mathcal{D})$, which can model $f(s \mid \mathcal{D},q)$ solely with MIPS, as well as close the gap in performance.

%!TEX root = ../acl2021.tex

\begin{figure*}[t]
\begin{center}
\includegraphics[height=7.2cm]{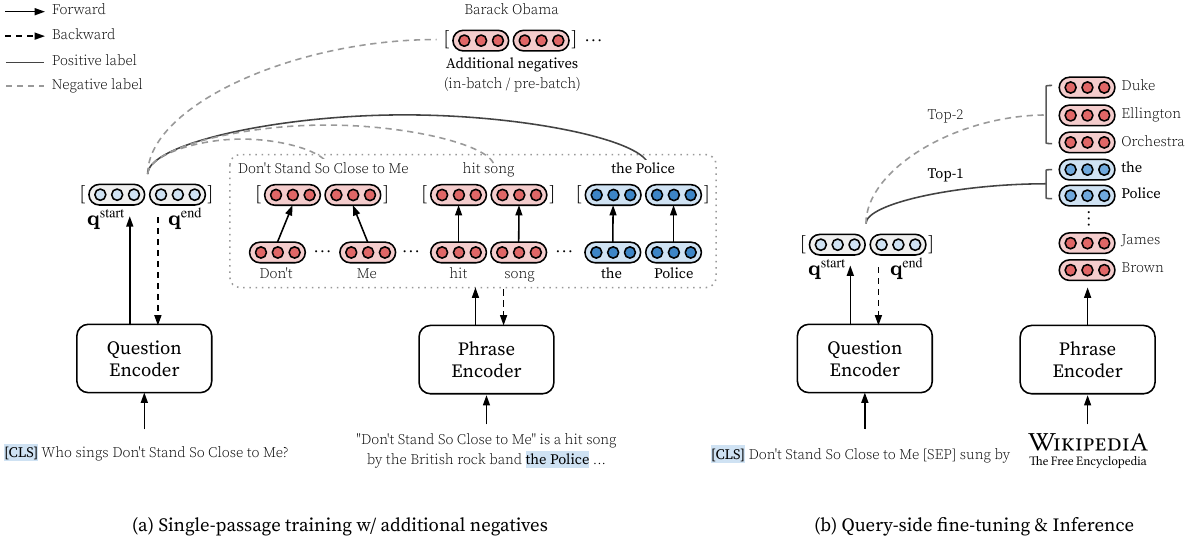}
\end{center}
\caption{An overview of \ours.  (a) We learn dense phrase representations in a single passage (\S\ref{sec:single-passage}) along with in-batch and pre-batch negatives (\S\ref{sec:inbatch}, \S\ref{sec:prebatch}).
%The question vector is represented by $\mathbf{q}^\text{start}$ and $\mathbf{q}^\text{end}$, which are trained to match the start and end positions of the gold phrase (e.g., `the Police').
%Note that the use of data augmentation and knowledge distillation is omitted in the figure.
(b) With the top-$k$ retrieved phrase representations from the entire text corpus (\S\ref{sec:indexing_and_search}), we further perform query-side fine-tuning to optimize the question encoder (\S\ref{sec:qsft}).
% , which can also adapt our model to new types of questions .
During inference, our model simply returns the top-1 prediction.
% \todo{Single-Passage $\Longrightarrow$ Single-passage, Query-Side Fine-Tuning $\Longrightarrow$ Query-side Fine-tuning}
%\todo{Suggestion: How about we use $\tilde{q}$ for both (c) and (d)? This is to emphasize that $\tilde{q}$ can be different from $q$ used in (a). You can even add a short sentence in the caption. }
}\label{fig:overview}
\end{figure*}

% Each stage is processed in the following order: (a) $\rightarrow$ (b) $\rightarrow$ (c) $\rightarrow$ (d).
% \danqi{Define $\wordset$ in caption.}\jinhyuk{added}
 % We also tackle the decomposability gap (\Cref{sec:piqa}) and the normalization issue (\Cref{sec:normalization}) in this stage.

% % \paragraph{Technical Challenges}
% Despite an appealing approach, phrase retrieval poses several key technical challenges.
% The first challenge is the decomposition constraint between question and phrase encoders as stated in Eq.~\eqref{eqn:decomp}, which brings a significant degradation of performance, compared to cross-attention models~\cite{devlin2019bert}.
% % While a similar problem is observed in learning dual encoders for passage representations~\citep{humeau2019poly,khattab2020colbert}, phrase representations are even more difficult to learn due to the fine-grained representations.
% The second challenge arises from the scale.
% Compared to 5 million documents, or 21 million passages as used in previous work, it is an inherently more challeging problem to retrieve from over 60 billion phrases.
% % This normalization problem is also implied in Equation~\eqref{eqn:decomp} that $E_{s}(s, \documentset)$ is defined over all the phrases in $\phraseset$.
% Lastly, it is computationally expensive to build billions of phrase representations at Wikipedia scale, making it prohibitive to update the phrase representations once they are obtained.
% % As a result, current phrase retrieval models often rely on their zero-shot ability~\citep{lee2020contextualized}.

%!TEX root = ../acl2021.tex

\section{{\ours}}

\subsection{Overview}

We introduce {\ours}, a phrase retrieval model that is built on fully dense representations.
Our goal is to learn a {phrase} encoder as well as a {question} encoder, so we can pre-index all the possible phrases in $\mathcal{D}$, and efficiently retrieve phrases for any question through MIPS at testing time. We outline our approach as follows:
\begin{itemize}[noitemsep]
\item
We first learn a high-quality phrase encoder and an (initial) question encoder from the supervision of reading comprehension tasks (\S\ref{sec:single-passage}), as well as incorporating effective negative sampling to better discriminate phrases at scale (\S\ref{sec:inbatch}, \S\ref{sec:prebatch}).
\item
Then, we fix the phrase encoder and encode all the phrases $s \in \phraseset$ and store the phrase indexing offline to enable efficient search (\S\ref{sec:indexing_and_search}).
\item
Finally, we introduce an additional strategy called query-side fine-tuning (\S\ref{sec:qsft}) by further updating the question encoder.\footnote{In this paper, we use the term \ti{question} and \ti{query} interchangeably as our question encoder can be naturally extended to ``unnatural'' queries.} We find this step to be very effective, as it can reduce the discrepancy between training (the first step) and inference, as well as support transfer learning to new domains.
\end{itemize}

Before we present the approach in detail, we first describe our base architecture below.

% As illustrated in Figure~\ref{fig:sample}, we first describe our query-agnostic model in a single-passage setting and address the decomposability gap (\Cref{sec:piqa}). Then we propose several normalization techniques for scaling phrase representations to the full collection of documents $\mathcal{D}$ (\Cref{sec:normalization}).
% Finally, we detail how we adapt our model for transfer learning~(\Cref{sec:qsft}), without the need for re-building of phrase representations at scale.

\subsection{Base Architecture}
\label{sec:base_model}
% first describe our base architecture, which
Our base architecture consists of a phrase encoder $E_s$ and a question encoder $E_q$. Given a passage $p = w_1, \ldots, w_m$, we denote all the phrases up to $L$ tokens as $\phraseinp$. Each phrase $s_k$ has start and end indicies $\ttt{start}(k)$ and $\ttt{end}(k)$ and the gold phrase is $s^* \in \phraseinp$.
%\footnote{In some reading comprehension datasets (e.g., SQuAD, Natural Questions), the gold phrase $s^*$ is given. While in some other datasets, only the gold answer $a^*$ is provided and we need to find a phrase $s$ that matches $\ttt{TEXT}(s) = a^*$. }
% $w_1, \ldots, w_m$
Following previous work on phrase or span representations~\citep{lee2017learning,seo2018phrase}, we first apply a pre-trained language model $\lm_p$ to obtain contextualized word representations for each passage token: $\mf{h}_1, \dots, \mf{h}_m \in \mathbb{R}^d$. Then, we can represent each phrase $s_k \in \phraseinp$ as the concatenation of corresponding start and end vectors:
\vspace{-0.5em}
\begin{equation}
E_s(s_k, p) = [\mf{h}_{\ttt{start}(k)}, \mf{h}_{\ttt{end}(k)}] \in \mathbb{R}^{2d}.
\end{equation}
% Using contextualized word representations to construct phrase representations has another great advantage that we can eventually reduce the storage of phrase representations to word representations. Therefore
A great advantage of this representation is that we eventually only need to index and store all the word vectors (we use $\wordset$ to denote all the words in $\documentset$), instead of all the phrases $\phraseset$, which is at least one magnitude order smaller.

Similarly, we need to learn a question encoder $E_q(\cdot)$ that maps a question $q = \tilde{w}_1, \ldots, \tilde{w}_{n}$ to a vector of the same dimension as $E_s(\cdot)$. Since the start and end representations of phrases are produced by the same language model, we use another two different pre-trained encoders $\lm_{q, \text{start}}$ and $\lm_{q, \text{end}}$ to differentiate the start and end positions. We apply $\lm_{q, \text{start}}$ and $\lm_{q, \text{end}}$ on $q$ separately and obtain representations $\mf{q}^\text{start}$ and $\mf{q}^\text{end}$ taken from the \texttt{[CLS]} token representations respectively. Finally, $E_q(\cdot)$ simply takes their concatenation:
\vspace{-0.5em}
\begin{equation}
E_q(q) = [\mf{q}^\text{start}, \mf{q}^\text{end}] \in \mathbb{R}^{2d}.
\end{equation}
% \vspace{-0.5em}
Note that we use pre-trained language models to initialize $\lm_p$, $\lm_{q, \text{start}}$ and $\lm_{q, \text{end}}$ and they are fine-tuned with the objectives that we will define later. In our pilot experiments, we found that SpanBERT~\citep{joshi2020spanbert} leads to superior performance compared to BERT~\cite{devlin2019bert}.
SpanBERT is designed to predict the information in the entire span from its two endpoints, therefore it is well suited for our phrase representations.
In our final model, we use SpanBERT-base-cased as our base LMs for $E_s$ and $E_q$, and hence $d=768$.\footnote{Our base model is largely inspired by DenSPI~\cite{seo2019real}, although we deviate from theirs as follows. (1)
%Instead of splitting a hidden vector from a pre-trained LM into four vectors,(start/end vectors and two vectors for calculating a coherency score),
We remove coherency scalars and don't split any vectors.
% We find that keeping the output dimension of pre-trained LMs is better utilizing their representational capacity.
(2) DenSPI uses a shared encoder for phrases and questions while we use 3 separate language models initialized from the same pre-trained model. (3) We use SpanBERT instead of BERT. } See Table~\ref{tab:piqa-ablation} for an ablation study.

% \subsection{Roadmap}

%  Previous models split
% We don't do any splitting of vectors and remove the use of coherency scalars.
%
% (2) Previous models use a shared encoder for phrases and questions.
% However, we use two different language models for representing questions. (3) We use SpanBERT instead of BERT.
%
%
% % In summary, we have three different LMs in total, which are initialized from the same pre-trained LM.
%  % we need two question representations $\mf{q}^\text{start}$ and $\mf{q}^\text{end}$
%
% We also summarize the differences of our base model from DenSPI~\citep{seo2019real} in~\Cref{apdx:compare-denspi}.% \jinhyuk{moved to appendix}

% \vspace{-0.1cm}
\section{Learning Phrase Representations}
\label{sec:learning_phrases}

% In this section, we start by learning query-agnostic phrase representations in a reading comprehension setting, in which a gold passage $p$ is given for a question-answer pair ($q, a^*$).
In this section, we start by learning dense phrase representations from the supervision of reading comprehension tasks, i.e., a single passage $p$ contains an answer $a^*$ to a question $q$.
Our goal is to learn strong dense representations of phrases for $s \in \phraseinp$, which can be retrieved by a dense representation of the question and serve as a direct answer (\S\ref{sec:single-passage}).
Then, we introduce two different negative sampling methods (\S\ref{sec:inbatch}, \S\ref{sec:prebatch}), which encourage the phrase representations to be better discriminated at the full Wikipedia scale.
See Figure~\ref{fig:overview} for an overview of \ours.
% Our goal is to build a strong reading comprehension model while enforcing the decomposability of query and phrase representations.
% In the following, we first describe our base model (\cref{sec:base_model}) and propose two new solutions to close the decomposability gap: tackling data sparsity via question generation (\cref{sec:data_sparsity}) and distillation from query-dependent models (\cref{sec:distillation}).

\subsection{Single-passage Training}\label{sec:single-passage}
To learn phrase representations in a single passage along with question representations, we first maximize the log-likelihood of the start and end positions of the gold phrase $s^*$ where $\ttt{TEXT}{(s^*)}=a^*$.
The training loss for predicting the start position of a phrase given a question is computed as:
% \vspace{-1.5em}
\begin{equation}\label{eqn:pstart}
\begin{split}
z_1^\text{start}, \dots, z_m^\text{start} &=  [\mf{h}_1^{\top}\mf{q}^\text{start}, \dots, \mf{h}_m^{\top}\mf{q}^\text{start}], \\
    P^\text{start} &= \textrm{softmax}(z_1^\text{start}, \dots, z_m^\text{start}), \\
    \mathcal{L}_\text{start} &=  -\log P^\text{start}_{\ttt{start}(s^*)}.
\end{split}
\end{equation}
We can define $\mathcal{L}_\text{end}$ in a similar way and the final loss for the single-passage training is
\begin{equation}\label{eqn:lsingle}
\begin{split}
\mathcal{L}_\text{single} = \frac{\mathcal{L}_\text{start} + \mathcal{L}_\text{end}}{2}.
\end{split}
\end{equation}
This essentially learns reading comprehension without any cross-attention between the passage and the question tokens, which fully decomposes phrase and question representations.

\begin{comment}
\begin{equation}
    \mathcal{L}_\text{single} = \frac{\mathcal{L}_\text{start} + \mathcal{L}_\text{end}}{2}.
\end{equation}

\paragraph{Differences from DenSPI}
We deviate from DenSPI in the following ways: (1) Previous models split a hidden vector from a pre-trained LM into four vectors (start \& end vectors and two vectors for calculating a coherency score).
We don't do any splitting of vectors and remove the use of coherency scalars.
We find that it is beneficial to keep the output dimension of pre-trained LMs for fully utilizing their representational capacity;
(2) Previous models use a shared encoder for phrases and questions.
However, we use two different language models for representing questions. (3) We use SpanBERT instead of BERT. See Table~\ref{tab:piqa-ablation} for an ablation study.
\end{comment}

% \subsection{Tackling Data Sparsity}\label{sec:data_sparsity}
\paragraph{Data augmentation}
Since the contextualized word representations $\mf{h}_1, \dots, \mf{h}_m$ are encoded in a query-agnostic way, they are always inferior to \ti{query-dependent} representations in cross-attention models~\citep{devlin2019bert},  where passages are fed along with the questions concatenated by a special token such as \ttt{[SEP]}.
% We hypothesize that the small number of answer annotations in each passage of reading comprehension datasets might bias the token representations to the small number of phrases, hence causing the performance gap.
We hypothesize that one key reason for the performance gap is that reading comprehension datasets only provide a few annotated questions in each passage, compared to the set of possible answer phrases. Learning from this supervision is not easy to differentiate similar phrases in one passage (e.g., $s^*=$ \textit{Charles, Prince of Wales} and another $s=$ \textit{Prince George} for a question $q=$ \textit{Who is next in line to be the monarch of England?}).
% For instance, each passage in the training set of Natural Questions~\cite{kwiatkowski2019natural} mostly has only one annotated question.

\begin{comment}
Suppose that we are given a passage with the following question-answer pair in the training set:
\begin{quote}
\small
$p=$ \textit{Queen Elizabeth II is the sovereign, and her heir apparent is her eldest son, \textbf{Charles, Prince of Wales}. (...) Third in line is Prince George, the eldest child of the Duke of Cambridge (...)} \\
$q=$ \textit{who is next in line to be the monarch of england} \\
$s^*=$ \textit{\textbf{Charles, Prince of Wales}}
\end{quote}
\vspace{-0.5em}
\noindent While cross-attention models only need to represent the passage focusing on ``who is Queen Elizabeth II's heir apparent,'' our phrase encoder should take all the other phrases into account, (e.g., $s'=$ \textit{Prince George}), because their representations will be re-used for other questions (e.g., $q'=$ \textit{who is the eldest child of the duke of cambridge}).
\end{comment}

Following this intuition, we propose to use a simple model to generate additional questions for data augmentation, based on a T5-large model~\cite{raffel2020exploring}.
% We use a T5-large model~\citep{raffel2020exploring} for generating questions.
To train the question generation model, we feed a passage $p$ with the gold answer $s^*$ highlighted by inserting surrounding special tags.
Then, the model is trained to maximize the log-likelihood of the question words of $q$.
After training, we extract all the named entities in each training passage as candidate answers and feed the passage $p$ with each candidate answer to generate questions.
We keep the question-answer pairs only when a cross-attention reading comprehension model\footnote{SpanBERT-large, 88.2 EM on SQuAD.} makes a correct prediction on the generated pair.
The remaining generated QA pairs $\{(\bar{q}_1, \bar{s}_1), (\bar{q}_2, \bar{s}_2), \ldots, (\bar{q}_r, \bar{s}_{r})\}$ are directly augmented to the original training set.
% Generated QA pairs can help learn phrase representations aligned with those of corresponding questions, instead of biased to few annotated questions.

% \subsection{Distillation}
% \label{sec:distillation}
% As query-dependent (QD) models with cross-attention are considered stronger models,
\paragraph{Distillation} We also propose improving the phrase representations by distilling knowledge from a cross-attention model~\citep{hinton2015distilling}.
We minimize the Kullback–Leibler divergence between the probability distribution from our phrase encoder and that from a standard SpanBERT-base QA model. The loss is computed as follows:
\begin{equation}\label{eqn:distill}
    \mathcal{L}_\text{distill} = \frac{ \text{KL}(P^\text{start} || P^\text{start}_{c}) + \text{KL}(P^\text{end} || P^\text{end}_{c})}{2},
\end{equation}
where $P^\text{start}$ (and $P^\text{end}$) is defined in Eq.~\eqref{eqn:pstart} and $P^\text{start}_c$ and $P^\text{end}_c$ denote the probability distributions used to predict the start and end positions of answers in the cross-attention model.
% \footnote{We use a standard SpanBERT-base QA model as the query-dependent model.}

% \section{Phrase Representations at Scale}
% \label{sec:normalization}

%!TEX root = ../acl2021.tex

\begin{figure}[t]
\begin{center}
\includegraphics[height=4.2cm]{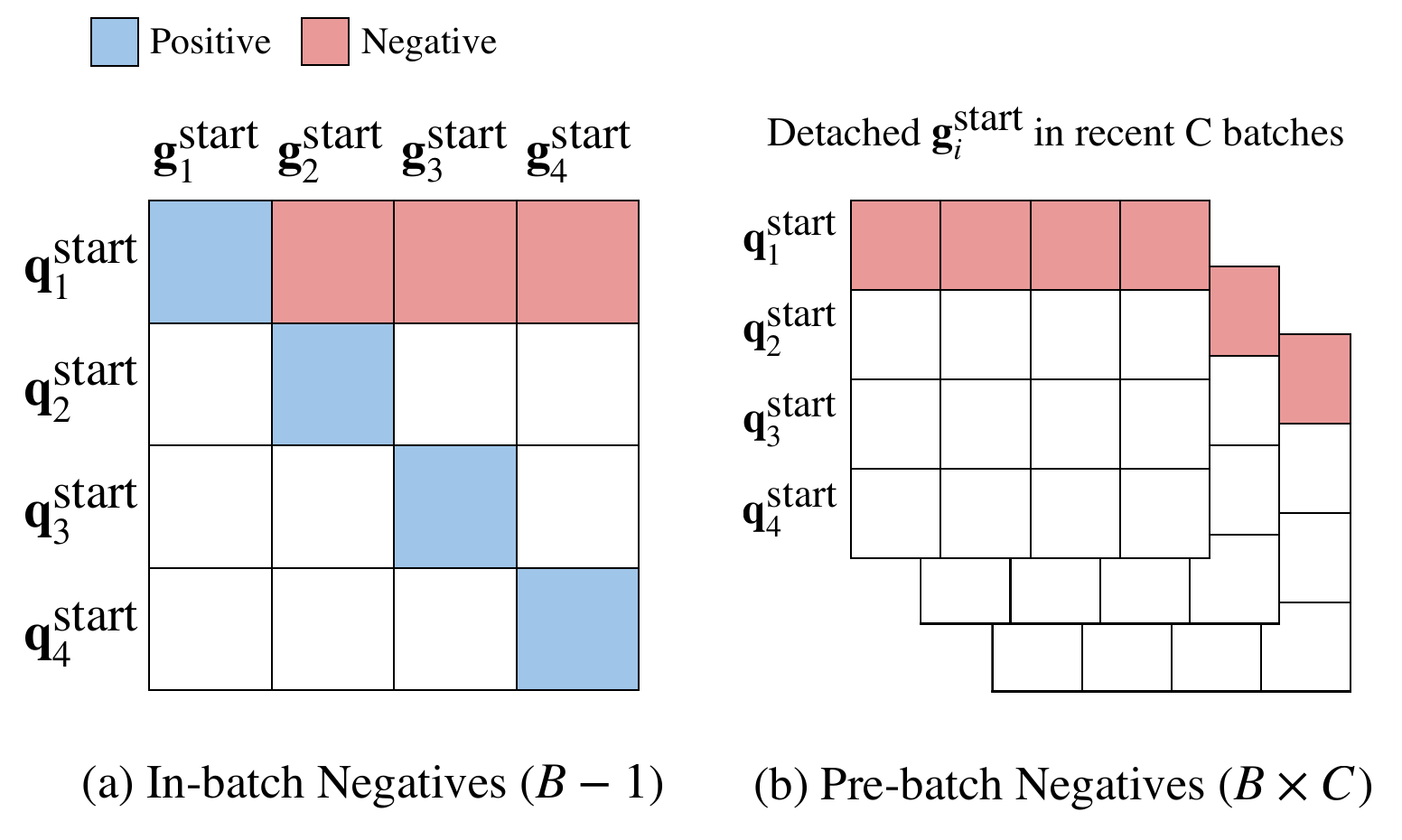}
\end{center}\vspace{-0.2cm}
\caption{Two types of negative samples for the first batch item ($\mathbf{q}_1^\text{start}$) in a mini-batch of size $B=4$ and $C=3$.  Note that the negative samples for the end representations ($\mathbf{q}_i^\text{end}$) are obtained in a similar manner. See \S\ref{sec:inbatch} and \S\ref{sec:prebatch} for more details.
% (a) In-batch negatives use gold phrase representations of other passages in the same mini-batch as negative samples. (b) Pre-batch negatives use detached gold phrase representations in recent $C$ mini-batches (e.g., $C=3$) as negative samples.
}\vspace{-0.3cm}\label{fig:batchneg}
\end{figure}

% We also denote the positive sample for each batch item in the current mini-batch.

\subsection{In-batch Negatives}\label{sec:inbatch}
Eventually, we need to build phrase representations for billions of phrases. Therefore, a bigger challenge is to incorporate more phrases as negatives so the representations can be better discriminated at a larger scale.
While \citet{seo2019real} simply sample two negative passages based on question similarity, we use in-batch negatives for our dense phrase representations, which has been shown to be effective in learning dense passage representations before~\cite{karpukhin2020dense}.

As shown in Figure~\ref{fig:batchneg} (a), for the $i$-th example in a mini-batch of size $B$, we denote the hidden representations of the gold start and end positions $\mf{h}_{\ttt{start}(s^*)}$ and $\mf{
h}_{\ttt{end}(s^*)}$ as $\mf{g}^{\text{start}}_i$ and $\mf{g}^{\text{end}}_i$, as well as the question representation as $[\mf{q}^{\text{start}}_i, \mf{q}^{\text{end}}_i]$. Let $\mf{G}^{\text{start}}, \mf{G}^{\text{end}}, \mf{Q}^{\text{start}}, \mf{Q}^{\text{end}}$ be the $B \times d$ matrices and each row corresponds to $\mf{g}^{\text{start}}_i, \mf{g}^{\
\text{end}}_i, \mf{q}^{\text{start}}_i, \mf{q}^{\text{end}}_i$ respectively. Basically, we can treat all the gold phrases from other passages  in the same mini-batch as negative examples.  We compute $\mf{S}^{\text{start}} = {\mf{Q}^{\text{start}}}{\mf{G}^{\text{start}}}^{\intercal}$ and $\mf{S}^{\text{end}}
 = {\mf{Q}^{\text{end}}}{\mf{G}^{\text{end}}}^{\intercal}$ and the $i$-th row of $\mf{S}^{\text{start}}$ and $\mf{S}^{\text{end}}$ return $B$ scores each, including a positive score and $B$$-1$ negative scores: $s^{\text{start}}_1, \ldots, s^{\text{start}}_B$ and $s^{\text{end}}_1, \ldots, s^{\text{end}}_B$. Similar to Eq.~\eqref{eqn:pstart}, we can compute the loss function for the $i$-th example as:\vspace{-0.5em}
\begin{equation}\label{eqn:inbatch}
\begin{split}
    P^\text{start\_ib}_{i} &= \textrm{softmax}(s_1^\text{start}, \dots, s_B^\text{start}), \\
    P^\text{end\_ib}_{i} &= \textrm{softmax}(s_1^\text{end}, \dots, s_B^\text{end}), \\
    \mathcal{L}_\text{neg} &= -\frac{\log P^\text{start\_ib}_{i}+\log P^\text{end\_ib}_{i}}{2},
\end{split}
\end{equation}
% where $P^\text{end\_ib}_{i}$ is defined similarly as $P^\text{start\_ib}_{i}$.
% Finally, the loss is summed over $B$ examples in the mini-batch.
We also attempted using non-gold phrases from other passages as negatives but did not find a meaningful improvement.

\subsection{Pre-batch Negatives}\label{sec:prebatch}
The in-batch negatives usually benefit from a large batch size~\citep{karpukhin2020dense}.
However, it is challenging to further increase batch sizes, as they are bounded by the size of GPU memory.
Next, we propose a novel negative sampling method called \ti{pre-batch negatives}, which can effectively utilize the representations from the preceding $C$ mini-batches (Figure~\ref{fig:batchneg} (b)).
In each iteration, we maintain a FIFO queue of $C$ mini-batches to cache phrase representations $\mf{G}^{\text{start}}$ and $\mf{G}^{\text{end}}$.  The cached phrase representations are then used as negative samples for the next iteration, providing $B \times C$ additional negative samples in total.\footnote{This approach is inspired by the momentum contrast idea proposed in unsupervised visual representation learning~\cite{he2020momentum}. Contrary to their approach, we have separate encoders for phrases and questions and back-propagate to both during training without a momentum update.}
% Figure~\ref{fig:batchneg} shows an illustrative example of in-batch and pre-batch negatives.
% \footnote{In their approach, the query and key encoders are shared. Only the query encoder is back-propagated and the key encoder is updated as a moving average. Interested readers are referred to their paper for details.}

These pre-batch negatives are used together with in-batch negatives and the training loss is the same as Eq.~\eqref{eqn:inbatch}, except that the gradients are \ti{not} back-propagated to the cached pre-batch negatives.
% In practice, we found that pre-batch negatives work well, once the phrase encoder is warmed up with in-batch negatives.
After warming up the model with in-batch negatives, we simply shift from in-batch negatives ($B - 1$ negatives) to in-batch and pre-batch negatives (hence a total number of $B \times C + B - 1$ negatives).
For simplicity, we use $\mathcal{L}_\text{neg}$ to denote the loss for both in-batch negatives and pre-batch negatives. % during training.
Since we do not retain the computational graph for pre-batch negatives, the memory consumption of pre-batch negatives is much more manageable while allowing an increase in the number of negative samples.
% Empirically, we found that using a large number of pre-batch negatives does not always help since the phrase representations can get easily outdated.

% \subsection{Optimization, Indexing and Search}\label{sec:optimization}
%With our loss terms defined previously,
\subsection{Training Objective} Finally, we optimize all the three losses together, on both annotated reading comprehension examples and generated questions from \S\ref{sec:single-passage}:
% We finally minimize the following loss function on a question answering dataset, together with the generated questions  from \Cref:
% \vspace{-0.4em}
\begin{equation}\label{eqn:aggregate}
    \mathcal{L} = \lambda_1 \mathcal{L}_\text{single} + \lambda_2 \mathcal{L}_\text{distill} + \lambda_3 \mathcal{L}_\text{neg},
\end{equation}
\noindent where $\lambda_1,\lambda_2,\lambda_3$ determine the importance of each loss term.
We found that $\lambda_1=1$, $\lambda_2=2$, and $\lambda_3=4$ works well in practice. See  Table~\ref{tab:piqa-ablation} and Table~\ref{tab:sod-qa} for an ablation study of different components.
% We use reading comprehension datasets (a gold passage $p$ is provided) such as Natural Questions to train the phrase and question encoders.

% \section{Query-side Fine-tuning \& Inference}
% \label{sec:qsft}

\section{Indexing and Search}
\label{sec:indexing_and_search}

\paragraph{Indexing}
After training the phrase encoder $E_s$, we need to encode all the phrases $\phraseset$ in the entire English Wikipedia $\documentset$ and store an index of the phrase dump.
We segment each document $d_i \in \documentset$ into a set of natural paragraphs, from which we obtain token representations for each paragraph using $E_s(\cdot)$.
Then, we build a phrase dump $\phrasedump = [\mf{h}_1, \dots, \mf{h}_{|\wordset|}]\in \mathbb{R}^{|\wordset| \times d}$ by stacking the token representations from all the paragraphs in $\documentset$.
Note that this process is computationally expensive and takes about hundreds of GPU hours with a large disk footprint.
To reduce the size of phrase dump, we follow and modify several techniques introduced in~\citet{seo2019real} (see \Cref{apdx:storage} for details).
After indexing, we can use two rows $i$ and $j$ of $\phrasedump$ to represent a dense phrase representation $[\mf{h}_i, \mf{h}_j]$. We use \texttt{faiss}~\citep{johnson2017billion} for building a MIPS index of $\phrasedump$.\footnote{We use \texttt{IVFSQ4} with 1M clusters and set n-probe to 256.}

\paragraph{Search}
For a given question $q$, we can find the answer $\hat{s}$ as follows:
\vspace{-0.2em}
\begin{equation}\label{eqn:formula}
\begin{split}
    \hat{s} &= \argmax_{s_{(i,j)}} E_s(s_{(i,j)}, \documentset) ^\top E_q(q), \\
    % &=  \argmax_{s_{(i,j)}} \mf{h}_i ^\top \mf{q}^\text{start} + \mf{h}_j ^\top \mf{q}^\text{end} \\
    &=  \argmax_{s_{(i,j)}} (\phrasedump \mf{q}^\text{start})_i + (\phrasedump \mf{q}^\text{end})_j,
\end{split}
\end{equation}
\noindent where $s_{(i,j)}$ denotes a phrase with start and end indices as $i$ and $j$ in the index $\phrasedump$.
We can compute the $\argmax$ of $\phrasedump \mf{q}^\text{start}$ and $\phrasedump \mf{q}^\text{end}$ efficiently by performing MIPS over $\phrasedump$ with $\mf{q}^\text{start}$ and $\mf{q}^\text{end}$.
In practice, we search for the top-$k$ start and top-$k$ end positions separately and perform a constrained search over their end and start positions respectively such that $1 \leq i \leq j < i+L \leq |\wordset|$.
% Since we share $\phrasedump$ for the start and end representations, $\mf{q}^\text{start}$ and $\mf{q}^\text{end}$ are also batched for MIPS to benefit from multi-threading.
% To avoid producing redundant answers, we only keep the best scoring phrases when the two phrases that have the same normalized string are retrieved from the same paragraph.

% that can be directly used for question answering.

% which can facilitate {transfer learning} on a new dataset,

\section{Query-side Fine-tuning}
\label{sec:qsft}
So far, we have created a phrase dump $\phrasedump$ that supports efficient MIPS search. In this section, we propose a novel method called {query-side fine-tuning} by only updating the question encoder $E_q$ to correctly retrieve a desired answer $a^*$ for a question $q$ given $\phrasedump$.
Formally speaking, we optimize the marginal log-likelihood of the gold answer $a^*$ for a question $q$, which resembles the weakly-supervised QA setting in previous work~\citep{lee2019latent,min2019discrete}.
For every question $q$, we retrieve top $k$ phrases and minimize the objective:
% The loss for query-side fine-tuning is computed as follows:
% \begin{equation}\label{eqn:qsft}
%     \mathcal{L}_\text{open} = -\log \sum_{\substack{s \in \tilde{\mathcal{S}}(q) \\ {\ttt{TEXT}(s) = a^*}}}\frac{\exp{\big(f(s|\documentset, q)\big)}}{\sum\limits_{s_i \in \tilde{\mathcal{S}}(q)} \exp{\big(f(s_i|\documentset,q)\big)}},
% \end{equation}
\begin{equation}\label{eqn:qsft}
\resizebox{1.0\columnwidth}{!}{
    $\mathcal{L}_\text{query} = -\log \frac{\sum_{\substack{s \in \tilde{\mathcal{S}}(q), {\ttt{TEXT}(s) = a^*}}} \exp{\big(f(s|\documentset, q)\big)}}{\sum_{s \in \tilde{\mathcal{S}}(q)} \exp{\big(f(s|\documentset, q)\big)}},$
}
\end{equation}
where $f(s|\documentset, q)$ is the score of the phrase $s$ (Eq.~\eqref{eqn:decomp}) and $\tilde{\mathcal{S}}(q)$ denotes the top $k$ phrases for $q$ (Eq.~\eqref{eqn:formula}).
In practice, we use $k=100$ for all the experiments.
% Note that only the parameters of the question encoder $E_q$ are updated.

There are several advantages for doing this: (1) we find that query-side fine-tuning can reduce the discrepancy between training and inference, and hence improve the final performance substantially (\S\ref{sec:qsft-ablation}). Even with effective negative sampling, the model only sees a small portion of passages compared to the full scale of $\documentset$ and this training objective can effectively fill in the gap. (2) This training strategy allows for transfer learning to unseen domains, without rebuilding the entire phrase index. More specifically, the model is able to quickly adapt to new QA tasks (e.g., WebQuestions) when the phrase dump is built using SQuAD or Natural Questions. We also find that this can transfers to non-QA tasks when the query is written in a different format.
In~\Cref{sec:slot_filling}, we show the possibility of directly using \ours~ for slot filling tasks by using a query such as \ti{(Michael Jackson, is a singer of, x)}. In this regard, we can view our model as a dense knowledge base that can be accessed by many different types of queries and it is able to return phrase-level knowledge efficiently.

 % (3) it also creates a possibility to adapt our DensePhrases to non-QA tasks

% (2) even for the QA datasets used to build $\phrasedump$ (SQuAD and NQ in our experiments), we also find that query-side fine-tuning can further improve performance because it can reduce the discrepancy between training and inference;
% it helps the model quickly adapt to new QA tasks \ti{without} re-building billions of phrase representations;\footnote{Previous work~\cite{lee2020contextualized} used $E_{q}$ directly for additional QA tasks such as TREC. }
% \input{tables/piqa}
% \input{tables/open_qa}

%!TEX root = ../acl2021.tex

\section{Experiments}
\label{sec:experiments}

\subsection{Setup}
\paragraph{Datasets.}
We use two {reading comprehension} datasets: SQuAD~\cite{rajpurkar2016squad} and Natural Questions (NQ)~\cite{kwiatkowski2019natural} to learn phrase representations, in which a single gold passage is provided for each question.
% For Natural Questions, we use the short answer as a ground truth answer $a^*$ and its long answer as a gold passage $p$ (Appendix~\ref{apdx:prepro}).
% We train \ours~on these two datasets independently, and report its performance with other query-agnostic models. % in the reading comprehension setting.
For the open-domain QA experiments, we evaluate our approach on five popular {open-domain QA} datasets: Natural Questions, WebQuestions (WQ)~\citep{berant2013semantic}, CuratedTREC (TREC)~\citep{baudivs2015modeling}, TriviaQA (TQA)~\citep{joshi2017triviaqa}, and SQuAD.
Note that we only use SQuAD and/or NQ to build the phrase index and perform query-side fine-tuning (\S\ref{sec:qsft}) for other datasets.
% Although many questions in SQuAD are context-dependent, we evaluate our model on SQuAD for the comparison with previous phrase retrieval models~\citep{seo2019real,lee2020contextualized}, which were mainly trained and evaluated on SQuAD.

We also evaluate our model on two {slot filling} tasks, to show how to adapt our {\ours} for other knowledge-intensive NLP tasks.
We focus on using two slot filling datasets from the KILT benchmark~\citep{petroni2020kilt}: T-REx~\citep{elsahar2018t} and zero-shot relation extraction~\citep{levy2017zero}.
Each query is provided in the form of ``\{subject entity\} \ttt{[SEP]} \{relation\}" and the answer is the object entity.
\Cref{apdx:prepro} provides the statistics of all the datasets.

\paragraph{Implementation details.}
% Our phrase and question encoders are trained on each training set with Eq.~\eqref{eqn:aggregate}.
% We use SQuAD to train our QG model.\footnote{The quality of generated questions from a QG model trained on Natural Questions is worse due to the ambiguity of information-seeking questions.}
We denote the training datasets used for reading comprehension (Eq.~\eqref{eqn:aggregate}) as $\traincorpus_\text{phrase}$.
For open-domain QA, we train two versions of phrase encoders, each of which are trained on $\traincorpus_\text{phrase}=\{\text{SQuAD}\}$ and $\{\text{NQ}, \text{SQuAD}\}$, respectively.
% Note that both DenSPI and DenSPI + Sparc are trained on $\traincorpus_\text{phrase}=\{\text{SQuAD}\}$.
We build the phrase dump $\phrasedump$ for the 2018-12-20 Wikipedia snapshot and perform query-side fine-tuning on each dataset using Eq.~\eqref{eqn:qsft}.
% While we use a single 48GB GPU (Quadro RTX 8000) for training the phrase encoders with Eq.~\eqref{eqn:aggregate}, query-side fine-tuning is relatively cheap and uses a single 12GB GPU (TITAN Xp).
For slot filling, we use the same phrase dump for open-domain QA, $\traincorpus_\text{phrase}$ $=\{\text{NQ}, \text{SQuAD}\}$  and perform query-side fine-tuning on randomly sampled 5K or 10K training examples to see how rapidly our model adapts to the new query types.
See~\Cref{apdx:hyper} for details on the hyperparameters and \Cref{apdx:complexity} for an analysis of computational cost.

%!TEX root = ../acl2021.tex

\begin{table}[t]
    \centering
    \resizebox{0.95\columnwidth}{!}{%
    \begin{tabular}{lcccc}
        \toprule
         \multirow{2}{*}{\textbf{Model}} & \multicolumn{2}{c}{\textbf{SQuAD}} & \multicolumn{2}{c}{\textbf{NQ (Long)}} \\\cmidrule{2-3} \cmidrule{4-5}
        & EM & F1 & EM & F1 \\
        \midrule

        \multicolumn{3}{l}{\textit{Query-Dependent}}\\\midrule
        % DrQA & Bi-LSTM & 69.5 & 78.8 & - & - \\
        BERT-base & 80.8 & 88.5 & 69.9 & 78.2 \\
        % BERT-large & 84.1 & 90.9 & - & - \\
        SpanBERT-base & 85.7 & 92.2 & 73.2 & 81.0 \\
        % SpanBERT-large & - & - & - & -  \\
        \midrule

        \multicolumn{3}{l}{\textit{Query-Agnostic}} \\\midrule
        % LSTM + SA + ELMo & Bi-LSTM & 52.7 & 62.7 & - & - \\
        % DrQA (No $f_{aligned}$ and $f_{exact\_match}$) & Bi-LSTM & - & 59.4 & - & - \\
        DilBERT~\citep{siblini2020delaying} & \underline{63.0} & \underline{72.0} & - & - \\
        DeFormer~\citep{cao2020deformer} & - & \underline{72.1} & - & - \\
        DenSPI$^{\dagger}$ & 73.6 & 81.7 & 68.2 & 76.1 \\
        % DenSPI & SpanBERT-large & 74.6 & 82.9 & - & - \\
        DenSPI + Sparc$^{\dagger}$ & 76.4 & 84.8 & - & - \\
        % DenSPI + Sparc & SpanBERT-large & 78.0 & 85.7 & - & - \\
        \ours~(ours) & \textbf{78.3} & \textbf{86.3} & \textbf{71.9} & \textbf{79.6} \\
        % \midrule

        % \multicolumn{3}{l}{\textit{Ablation of \ours}} \\\midrule
        % \quad w/o Question Generation & 73.0 & 81.7 & (70.3) & (78.0) \\
        % \quad w/o Distillation & 76.3 & 84.4 & 68.1 & 75.9 \\
        % \quad SpanBERT => BERT & - & - & - & - \\

        \bottomrule
    \end{tabular}
    }\vspace{-0.0cm}
    \caption{Reading comprehension results, evaluated on the development sets of SQuAD and Natural Questions. Underlined numbers are estimated from the figures from the original papers. $^{\dagger}$: BERT-large model. }\label{tab:pi-qa}\vspace{-0.5cm}
\end{table}

%!TEX root = ../acl2021.tex

\begin{table*}[t]
    \centering
    \resizebox{1.95\columnwidth}{!}{%
    \begin{tabular}{llccccc}
        \toprule
        \textbf{Model} & &\textbf{NQ} & \textbf{WQ} & \textbf{TREC} & \textbf{TQA} &  \textbf{SQuAD}\\
         \midrule

        \textit{Retriever-reader} & $\traincorpus_\text{retr}$: (Pre-)Training \\ \midrule
        DrQA~\citep{chen2017reading} & - & - & 20.7 & 25.4 & - & 29.8 \\
        BERT + BM25~\citep{lee2019latent}  & - & 26.5 & 17.7 & 21.3 & 47.1 & \tf{33.2} \\
        ORQA~\citep{lee2019latent} & \{Wiki.\}$^\dagger$ & 33.3 & 36.4 & 30.1 & 45.0 & 20.2 \\
        REALM$_\text{News}$~\citep{guu2020realm} & \{Wiki., CC-News\}$^\dagger$& 40.4 & 40.7 & 42.9 & - & - \\
        DPR-multi~\citep{karpukhin2020dense}  & \{NQ, WQ, TREC, TQA\} & \tf{41.5} & \tf{42.4} & \tf{49.4} & \tf{56.8} & 24.1 \\
        % RAG-Sequence~\citep{lewis2020retrieval} & 44.5 & 56.1 & \textbf{45.2} & \textbf{52.2} & - \\
        % FID (large)~\citep{izacard2020leveraging} & \textbf{51.4} & \textbf{67.6} & - & - & \textbf{56.7} \\
        % Distill Retriever-(T5-large)~\citep{anonymous2021distilling} & - & \textbf{52.5} & - & - & - & - & - & \textbf{71.5} & - & - \\
        \midrule

        % \textit{Reader Only} \\\midrule
        % T5-11B~\citep{roberts2020much} & 32.6 & 42.3 & 37.2 & - & -  \\
        % T5-11B + SSM~\citep{roberts2020much} & 34.8 & 51.0 & 40.8 & - & -  \\
        % \midrule

        \textit{Phrase retrieval} & $\traincorpus_\text{phrase}$: Training \\\midrule
        DenSPI~\citep{seo2019real} & \{SQuAD\} & 8.1$^{*}$ & 11.1$^{*}$ & 31.6$^{*}$ & 30.7$^{*}$ & 36.2 \\ % 0.71
        DenSPI + Sparc~\citep{lee2020contextualized} & \{SQuAD\} & 14.5$^{*}$ & 17.3$^{*}$ & 35.7$^{*}$ & 34.4$^{*}$ & \textbf{40.7}  \\
        DenSPI + Sparc~\citep{lee2020contextualized} & \{NQ, SQuAD\} & 16.5 & - & - & - & -  \\
        % DenSPI + Sparc (w/ QS Fine-Tune) & 26.2 & - & 33.7 & 42.7 & -  \\ % 0.78
        \ours~(ours) & \{SQuAD\}& 31.2 & 36.3 & 50.3 & \textbf{53.6} & 39.4 \\
        % \ours~(NQ) & \tf{40.9} & 49.2 & 37.1 & 49.7 & 25.7 \\
        \ours~(ours) & \{NQ, SQuAD\}&  \textbf{40.9} & \textbf{37.5} & \textbf{51.0} & 50.7 & 38.0 \\
        \bottomrule
    \end{tabular}
    }\vspace{-.1cm}
    \caption{Open-domain QA results. We report exact match (EM) on the test sets. We also show the additional training or pre-training datasets for learning the retriever models ($\traincorpus_\text{retr}$) and creating the phrase dump ($\traincorpus_\text{phrase}$). $^*$: no supervision using target training data (zero-shot). $^\dagger$: unlabeled data used for extra pre-training.
    }\vspace{-0.3cm}
    \label{tab:od-qa}
\end{table*}

\begin{table}[t]
    \centering
    \resizebox{0.95\columnwidth}{!}{%
    \begin{tabular}{lcccc}
        \toprule
        \multirow{2}{*}{\textbf{Model}}& \multicolumn{2}{c}{\textbf{T-REx}} & \multicolumn{2}{c}{\textbf{ZsRE}} \\\cmidrule{2-3} \cmidrule{4-5}
        & Acc & F1 & Acc & F1 \\
        \midrule

        % BART & 0.00 & 0.00 & 9.14 & 12.21 \\
        % T5 & 0.00 & 0.00& 9.02 & 13.52 \\
        DPR + BERT & - & - & 4.47 & 27.09 \\
        DPR + BART & 11.12 & 11.41 & 18.91 & 20.32 \\
        RAG & 23.12 & 23.94 & 36.83 & 39.91 \\
        \midrule
        % \ours~(5K) & 52.84 & 59.99 & \textbf{45.75} & \textbf{53.68} \\
        % \ours~(10K) & 55.16 & 61.44 & 43.80 & 52.75 \\
        \ours$^{5\text{K}}$ & 25.32 & 29.76 & 40.39 & 45.89 \\
        \ours$^{10\text{K}}$ & \textbf{27.84} & \textbf{32.34} & \textbf{41.34} & \textbf{46.79} \\
        \bottomrule
    \end{tabular}
    }\vspace{-0.1cm}
    \caption{Slot filling results on the test sets of T-REx and Zero shot RE (ZsRE) in the KILT benchmark.
    We report KILT-AC and KILT-F1 (denoted as \ti{Acc} and \ti{F1} in the table), which consider both span-level accuracy and correct retrieval of evidence documents.
    % We consider two settings, in which we use 5K and 10K training examples respectively.
    }\vspace{-0.5cm}\label{tab:slot_filling}
\end{table}

\subsection{Experiments: Question Answering}\label{sec:openqa_result}
\paragraph{Reading comprehension.}\label{sec:rc_result}
In order to show the effectiveness of our phrase representations, we first evaluate our model in the reading comprehension setting for SQuAD and NQ and report its performance with other query-agnostic models (Eq.~\eqref{eqn:aggregate} without query-side fine-tuning). This problem was originally formulated by \newcite{seo2018phrase} as the phrase-indexed question answering (PIQA) task.

Compared to previous query-agnostic models, our model achieves the best performance of 78.3 EM on SQuAD by improving the previous phrase retrieval model (DenSPI) by $4.7\%$ (Table~\ref{tab:pi-qa}). Although it is still behind cross-attention models, the gap has been greatly reduced and serves as a strong starting point for the open-domain QA model.

\paragraph{Open-domain QA.}
Experimental results on open-domain QA are summarized in Table~\ref{tab:od-qa}.
Without any sparse representations, \ours~outperforms previous phrase retrieval models by a large margin and achieves a $15\%$--$25\%$ absolute improvement on all datasets except SQuAD.
%As previous models only used SQuAD to train the phrase model and perform zero-shot prediction on other datasets, we add one more experiment training the model of \newcite{lee2020contextualized} on $\traincorpus_\text{phrase}=\{\text{NQ}, \text{SQuAD}\}$ for a fair comparison.
Training the model of \newcite{lee2020contextualized} on $\traincorpus_\text{phrase}=\{\text{NQ}, \text{SQuAD}\}$ only increases the result from 14.5\% to 16.5\% on NQ, demonstrating that it does not suffice to simply add more datasets for training phrase representations.
Our performance is also competitive with recent retriever-reader models~\cite{karpukhin2020dense}, while running much faster during inference (Table~\ref{tab:category}).
% \footnote{We can also consider using distantly-supervised examples of TriviaQA, WebQuestions and TREC for training our phrase representations, as the DPR-multi model did, and we leave it to future work.}
% Finally, we find that using $\traincorpus_\text{phrase} = \{\text{SQuAD}\}$ or $\{\text{NQ, SQuAD}\}$ doesn't make much difference after query-side fine-tuning in most datasets, except that including NQ in $\traincorpus_\text{phrase}$ brings a large gain of 9.7\% on the NQ evaluation.
% \danqi{I think we need to discuss the SQuAD results a bit.}

\subsection{Experiments: Slot Filling}\label{sec:slot_filling}
Table~\ref{tab:slot_filling} summarizes the results on the two slot filling datasets, along with the baseline scores provided by~\citet{petroni2020kilt}.
The only extractive baseline is DPR + BERT, which performs poorly in zero-shot relation extraction.
On the other hand, our model achieves competitive performance on all datasets and achieves state-of-the-art performance on two datasets using only 5K training examples.% (less than 5\% of the training data).
% This showcases how \ours~can be easily leveraged for knowledge-intensive NLP tasks.

%!TEX root = ../acl2021.tex

\section{Analysis}
\label{sec:analysis}

\paragraph{Ablation of phrase representations.}
Table~\ref{tab:piqa-ablation} shows the ablation result of our model on SQuAD.
% We observe that not sharing phrase and question encoders ({Share} = \xmark) and using the full output dimension ({Split} = \xmark) together improves the performance by 2\%.
% Using a stronger pre-trained LM SpanBERT leads to another 1.3\% improvement.
Upon our choice of architecture, augmenting training set with generated questions ({QG} = \cmark) and performing distillation from cross-attention models ({Distill} = \cmark) improve performance up to EM = 78.3.
We attempted adding the generated questions to the training of the SpanBERT-QA model but find a 0.3\% improvement, which validates that data sparsity is a bottleneck for query-agnostic models.

%!TEX root = ../acl2021.tex

\begin{table}[t]
    \centering
    \resizebox{0.95\columnwidth}{!}{%
    \begin{tabular}{lcccccc}
        \toprule
         \textbf{Model} & $\lm$ &
         {{Share}} & {{Split}} & {{QG}} &  {{Distill}} &  {{EM}} \\
        % \midrule
        % BERT-QA & Sb. & - & - & \xmark & - & 85.7 \\
        % & Sb. & - & - & \cmark & - & 86.0 \\

        \midrule
        DenSPI & Bb. & \cmark & \cmark & \xmark & \xmark & 70.2 \\
        & Sb. & \cmark & \cmark & \xmark & \xmark & 68.5 \\
        & Bl. & \cmark & \cmark & \xmark & \xmark & 73.6 \\
        \midrule
        Dense & Bb. & \cmark & \xmark  & \xmark & \xmark & 70.2 \\
        Phrases & Bb. & \xmark & \xmark  & \xmark & \xmark & 71.9 \\
        & Sb. & \xmark & \xmark  & \xmark & \xmark & 73.2 \\
        & Sb. & \xmark & \xmark  & \cmark & \xmark & 76.3 \\
        & Sb. & \xmark & \xmark  & \cmark & \cmark & \textbf{78.3} \\
        % \midrule

        \bottomrule
    \end{tabular}
    }
    \caption{Ablation of \ours~on the development set of SQuAD. Bb: BERT-base, Sb: SpanBERT-base, Bl: BERT-large. Share: whether question and phrase encoders are shared or not. Split: whether the full hidden vectors are kept or split into start and end vectors. QG: question generation (\S\ref{sec:single-passage}). Distill: distillation (Eq.\eqref{eqn:distill}). DenSPI~\cite{seo2019real} also included a coherency scalar and see their paper for more details.}\label{tab:piqa-ablation}\vspace{-0.4cm}
\end{table}

%!TEX root = ../acl2021.tex

\begin{table}[t]
    \centering
    \resizebox{0.90\columnwidth}{!}{%
    \begin{tabular}{lcccc}
        \toprule
        {Type} & $B$ & $C$ &
        $\documentset = \{p\}$ & $\documentset = \documentset_\text{small}$ \\
        \midrule
        None & 48 & - & 70.4 & 35.3 \\
        \midrule
        + In-batch & 48 & - & 70.5 & 52.4 \\ % sbcd_nq_inb48_new
        & 84 & - & 70.3 & 54.2 \\ % sbcd_nq_ftinb84_kl_x8
        \midrule
        + Pre-batch & 84 & 1 & 71.6 & 59.8 \\ % sbcd_nqqg_ginb84_x8_s192_pinb1_fix_re
        & 84 & 2 & \textbf{71.9} & \textbf{60.4} \\ % sbcd_nqqg_ginb84_x4_s192_pinb2_fix_re
        & 84 & 4 & 71.2 & 59.8 \\ % sbcd_nqqg_ginb84_x4_s192_pinb4_fix_re
        \bottomrule
    \end{tabular}
    }
    \caption{Effect of in-batch negatives and pre-batch negatives on the development set of Natural Questions. $B$: batch size. $C$: number of preceding mini-batches used in pre-batch negatives. $\documentset_\text{small}$: all the gold passages in the development set of NQ. $\{p\}$: single passage.
    %\todo{The results of Table 6 need to be discussed more in the text.}
    }\label{tab:sod-qa} \vspace{-0.3cm}
\end{table}

\paragraph{Effect of batch negatives.}\label{sec:semi_od}
We further evaluate the effectiveness of various negative sampling methods introduced in~\S\ref{sec:inbatch} and \S\ref{sec:prebatch}.
Since it is computationally expensive to test each setting at the full Wikipedia scale, we use a smaller text corpus $\documentset_\text{small}$ of all the gold passages in the development sets of Natural Questions, for the ablation study.
% We also report performance in the reading comprehension setting so $\documentset = \{p\}$ only consists of a gold passage.
Empirically, we find that results are generally well correlated when we gradually increase the size of $|\mathcal{D}|$.
% and we encourage interested readers to experiment with these settings for model development. % The results are summarized in Table~\ref{tab:sod-qa}.
As shown in Table~\ref{tab:sod-qa}, both in-batch and pre-batch negatives bring substantial improvements.
While using a larger batch size ($B=84$) is beneficial for in-batch negatives, the number of preceding batches in pre-batch negatives is optimal when $C=2$.
Surprisingly, the pre-batch negatives also improve the performance when $\documentset = \{p\}$.

\paragraph{Effect of query-side fine-tuning.}~\label{sec:qsft-ablation}
% The query-side fine-tuning further trains the question encoder $E_q$ with the phrase dump $\phrasedump$ and enables us to fine-tune \ours~on different types of questions.
% We use three different phrase encoders, each of which is trained on a different training dataset $\traincorpus_\text{phrase}$.
We summarize the effect of query-side fine-tuning in Table~\ref{tab:qsft-ablation}.
For the datasets that were not used for training the phrase encoders (TQA, WQ, TREC), we observe a 15\% to 20\% improvement after query-side fine-tuning. Even for the datasets that have been used (NQ, SQuAD), it leads to significant improvements (e.g., 32.6\%$\rightarrow$40.9\% on NQ for $\traincorpus_\text{phrase}$ = \{NQ\}) and it clearly demonstrates it can effectively reduce the discrepancy between training and inference.

% \input{tables/qsft_ablation}

%!TEX root = ../acl2021.tex

\section{Related Work}
\label{sec:related_work}

Learning effective dense representations of words is a long-standing goal in NLP~\citep{bengio2003neural,collobert2011natural,mikolov2013distributed,peters2018deep,devlin2019bert}.
% including the recent advances in contextualized word representations~\citep{peters2018deep,devlin2019bert,liu2019roberta}.
Beyond words, dense representations of many different granularities of text such as sentences~\citep{le2014distributed,kiros2015skip} or documents~\citep{yih2011learning} have been explored.
While dense phrase representations have been also studied for statistical machine translation~\citep{cho2014learning} or syntactic parsing~\citep{socher2010learning}, our work focuses on learning dense phrase representations for QA and any other knowledge-intensive tasks where phrases can be easily retrieved by performing MIPS.

This type of dense retrieval has been also studied for sentence and passage retrieval~\citep{humeau2019poly,karpukhin2020dense} (see~\citealp{lin2020pretrained} for recent advances in dense retrieval).
While DensePhrases is explicitly designed to retrieve phrases that can be used as an answer to given queries, retrieving phrases also naturally entails retrieving larger units of text, provided the datastore maintains the mapping between each phrase and the sentence and passage in which it occurs.
% \footnote{In practice, DensePhrases always maintains the mapping table between phrases and their original passages and returns predicted phrases along with their original passages.}
% We leave the study of such connection between different granularities as future work.

%!TEX root = ../main.tex

\begin{table}[t]
    \centering
    \resizebox{0.9\columnwidth}{!}{%
    \begin{tabular}{ccccccc}
        \toprule
         {QS} &
        {NQ} & {WQ} & {TREC} & {TQA} & {SQuAD} \\
        \midrule
        \multicolumn{6}{c}{{$\traincorpus_\text{phrase}$} = \{SQuAD\}} \\
        \midrule
        \xmark & 12.3 & 11.8 & 36.9 & 34.6  & 35.5 \\
        \cmark & 31.2 & 36.3 & 50.3 & \textbf{53.6} & \textbf{39.4} \\
        \midrule
        \multicolumn{6}{c}{{$\traincorpus_\text{phrase}$} = \{NQ\}} \\
        \midrule
        \xmark & 32.6 & 21.1 & 32.3 & 32.4 & 20.7 \\
        \cmark & \tf{40.9} & 37.1 & 49.7 & 49.2 & 25.7 \\
        \midrule
        \multicolumn{6}{c}{{$\traincorpus_\text{phrase}$} = \{NQ, SQuAD\}} \\
        \midrule
         \xmark & 28.9 & 18.9 & 34.9 & 31.9 & 33.2 \\
          \cmark & \textbf{40.9} & \textbf{37.5} & \textbf{51.0} & 50.7 & 38.0 \\
        \bottomrule
    \end{tabular}
    }
    \caption{Effect of query-side fine-tuning in \ours~on each test set. We report EM of each model before ({QS} = \xmark) and after ({QS} = \cmark) the query-side fine-tuning.}\label{tab:qsft-ablation} \vspace{-0.3cm}
\end{table}

%!TEX root = ../acl2021.tex

\section{Conclusion}
\label{sec:conclusion}

In this study, we show that we can learn dense representations of phrases at the Wikipedia scale, which are readily retrievable for open-domain QA and other knowledge-intensive NLP tasks.
We learn both phrase and question encoders from the supervision of reading comprehension tasks and introduce two batch-negative techniques to better discriminate phrases at scale.
We also introduce query-side fine-tuning that adapts our model to different types of queries.
% with a single 12GB GPU.
We achieve strong performance on five popular open-domain QA datasets, while reducing the storage footprint and improving latency significantly.
We also achieve strong performance on two slot filling datasets using only a small number of training examples, showing the possibility of utilizing our {\ours} as a knowledge base.

\vspace{1em}
\section*{Acknowledgments}
We thank Sewon Min, Hyunjae Kim, Gyuwan Kim, Jungsoo Park, Zexuan Zhong, Dan Friedman, Chris Sciavolino for providing valuable comments and feedback.
This research was supported by a grant of the Korea Health Technology R\&D Project through the Korea Health Industry Development Institute (KHIDI), funded by the Ministry of Health \& Welfare, Republic of Korea (grant number: HR20C0021) and National Research Foundation of Korea (NRF-2020R1A2C3010638).
It was also partly supported by the James Mi *91 Research Innovation Fund for Data Science and an Amazon Research Award.
%!TEX root = ../main.tex
\clearpage

\section*{Ethical Considerations}\label{sec:ethics}
Our work builds on standard reading comprehension datasets such as SQuAD to build phrase representations.
SQuAD, in particular, is created from a small number of Wikipedia articles sampled from top-10,000 most popular articles (measured by PageRanks), hence some of our models trained only on SQuAD could be easily biased towards the small number of topics that SQuAD contains.
We hope that excluding such datasets during training or inventing an alternative pre-training procedure for learning phrase representations could mitigate this problem.
Although most of our efforts have been made to reduce the computational complexity of previous phrase retrieval models (further detailed in~\Cref{apdx:complexity,apdx:storage}), leveraging our phrase retrieval model as a knowledge base will inevitably increase the minimum requirement for the additional experiments. We plan to apply vector quantization techniques to reduce the additional cost of using our model as a KB.

%\clearpage
\bibliographystyle{acl_natbib}
\bibliography{acl2021}

\clearpage
\appendix
\setcounter{table}{0}
\renewcommand{\thetable}{\Alph{section}.\arabic{table}}

\section{Computational Cost}~\label{apdx:complexity}
We describe the resources and time spent during inference (Table~\ref{tab:category} and \ref{tab:complexity}) and indexing (Table~\ref{tab:complexity}).
With our limited GPU resources (24GB $\times$ 4), it takes about 20 hours for indexing the entire phrase representations.
We also largely reduced the storage from 1,547GB to 320GB by (1) removing sparse representations and (2) using our sharing and split strategy.
See Appendix~\ref{apdx:storage} for the details on the reduction of storage footprint and Appendix~\ref{apdx:server} for the specification of our server for the benchmark.

%!TEX root = ../main.tex

\begin{table}[h]
    \centering
    \resizebox{1.0\columnwidth}{!}{%
    \begin{tabular}{llll}
        \toprule

        % \textbf{Training} & % \multicolumn{1}{l}{Resources} & Batch Size \\ \midrule
        % DPR & GPU (32Gb) $\times$ 8 & 256 \\
        % REALM & \multicolumn{1}{l}{TPU $\times$ 64} & 512 \\
        % DenSPI + Sparc & \multicolumn{1}{l}{GPU (24Gb) $\times$ 1} & 24 \\
        % \ours & \multicolumn{1}{l}{GPU (24Gb) $\times$ 1} & 84 $\times$ 2 \\
        % \midrule

        \textbf{Indexing} & Resources & Storage & Time \\ \midrule
        DPR & 32GB GPU $\times$ 8 & 76GB & 17h \\
        DenSPI + Sparc & 24GB GPU $\times$ 4 & 1,547GB & 85h \\% 75 + 10 \\
        \ours & 24GB GPU $\times$ 4 & 320GB & 20h \\ % 14 + 6
        \midrule
        % REALM & TPU $\times$ 16 & 21GB & - \\

        \textbf{Inference} & \multicolumn{1}{l}{RAM / GPU} & \multicolumn{2}{l}{\#Q/sec (\gpu{GPU}, \cpu{CPU})} \\ \midrule
        DPR & 86GB / 17GB & \multicolumn{2}{c}{\gpu{0.9}, \cpu{0.04}} \\
        DenSPI + Sparc & 27GB / 2GB & \multicolumn{2}{c}{\gpu{2.1}, \cpu{1.7}} \\
        \ours & 12GB / 2GB & \multicolumn{2}{c}{\gpu{20.6}, \cpu{13.6}} \\
        % REALM & \multicolumn{1}{l}{GPU (12GB) $\times$ 1} & - \\
        \bottomrule
    \end{tabular}
    }
    \caption{Complexity analysis of three open-domain QA models during indexing and inference. For inference, we also report the minimum requirement of RAM and GPU memory for running each model with \gpu{GPU}.
    For computing \#Q/s for \cpu{CPU}, we do not use GPUs but load all models on the RAM.
    %\jinhyuk{difficult to match training setups. pre-training? retriever? reader? qs fine-tuning? Maybe delete?}
    % \jinhyuk{maybe we can talk about CPU/GPU memory (\ours~consumes much smaller CPU/GPU memory than DPR)}
    }\vspace{-0.2cm}
    \label{tab:complexity}
\end{table}

\section{Server Specifications for Benchmark}\label{apdx:server}
To compare the complexity of open-domain QA models, we install all models in Table~\ref{tab:category} on the same server using their public open-source code.
Our server has the following specifications:

\begin{table}[h]
    \centering
    \resizebox{0.8\columnwidth}{!}{%
    \begin{tabular}{llll}
        \toprule
        Hardware \\ \midrule
        Intel Xeon CPU E5-2630 v4 @ 2.20GHz \\
        128GB RAM \\
        12GB GPU (TITAN Xp) $\times$ 2\\
        2TB 970 EVO Plus NVMe M.2 SSD $\times$ 1\\
        \bottomrule
    \end{tabular}
    }
    \caption{Server specification for the benchmark
    }
    \label{tab:server}
\end{table}

For DPR, due to its large memory consumption, we use a similar server with a 24GB GPU (TITAN RTX).
For all models, we use 1,000 randomly sampled questions from the Natural Questions development set for the speed benchmark and measure \#Q/sec.
We set the batch size to 64 for all models except BERTSerini, ORQA and REALM, which do not allow a batch size of more than 1 in their open-source implementations.
\#Q/sec for DPR includes retrieving passages and running a reader model and the batch size for the reader model is set to 8 to fit in the 24GB GPU (retriever batch size is still 64).
For other hyperparameters, we use the default settings of each model.
We also exclude the time and the number of questions in the first five iterations for warming up each model.
Note that despite our effort to match the environment of each model, their latency can be affected by various different settings in their implementations such as the choice of library (PyTorch vs. Tensorflow).

% \section{Differences from DenSPI}\label{apdx:compare-denspi}
% We deviate from DenSPI in the following ways: (1) Previous models split a hidden vector from a pre-trained LM into four vectors (start \& end vectors and two vectors for calculating a coherency score).
% We don't do any splitting of vectors and remove the use of coherency scalars.
% We find that it is beneficial to keep the output dimension of pre-trained LMs for fully utilizing their representational capacity;
% (2) Previous models use a shared encoder for phrases and questions.
% However, we use two different language models for representing questions. (3) We use SpanBERT instead of BERT.
% See Table~\ref{tab:piqa-ablation} for an ablation study.

\section{Data Statistics and Pre-processing}\label{apdx:prepro}
%!TEX root = ../main.tex

\begin{table}[t]
\label{table:dataset}
\begin{center}
\centering
\resizebox{0.9\columnwidth}{!}{%
\begin{tabular}{lrrr}
\toprule
\multicolumn{1}{l}{\bf Dataset}  & {\bf Train} & {\bf Dev} & {\bf Test}\\
\midrule
Natural Questions & 79,168 & 8,757 & 3,610 \\
WebQuestions & 3,417 & 361 & 2,032 \\
CuratedTrec & 1,353 & 133 & 694 \\
TriviaQA & 78,785 & 8,837 & 11,313 \\
SQuAD &  78,713 & 8,886 & 10,570 \\
\midrule

T-REx & 2,284,168 & 5,000 & 5,000 \\
Zero-Shot RE & 147,909 & 3,724 & 4,966 \\
\bottomrule
\end{tabular}
}
\end{center}
\caption{Statistics of five open-domain QA datasets and two slot filling datasets. We follow the same splits in open-domain QA for the two reading comprehension datasets (SQuAD and Natural Questions).% \jinhyuk{RAG seems to use all 2.2M and 147K}
}\vspace{-0.3cm}\label{table:openqa-data}
\end{table}

In Table~\ref{table:openqa-data}, we show the statistics of five open-domain QA datasets and two slot filling datasets.
Pre-processed open-domain QA datasets are provided by~\citet{chen2017reading} except Natural Questions and TriviaQA.
We use a version of Natural Questions and TriviaQA provided by~\citet{min2019discrete,lee2019latent}, which are pre-processed for the open-domain QA setting.
Slot filling datasets are provided by~\citet{petroni2020kilt}.
We use two reading comprehension datasets (SQuAD and Natural Questions) for training our model on Eq.~\eqref{eqn:aggregate}.
For SQuAD, we use the original dataset provided by the authors~\citep{rajpurkar2016squad}.
For Natural Questions~\citep{kwiatkowski2019natural}, we use the pre-processed version provided by~\citet{asai2019learning}.\footnote{\url{https://github.com/AkariAsai/learning\_to\_retrieve\_reasoning\_paths}}
We use the short answer as a ground truth answer $a^*$ and its long answer as a gold passage $p$.
We also match the gold passages in Natural Questions to the paragraphs in Wikipedia whenever possible.
Since we want to check the performance changes of our model with the growing number of tokens, we follow the same split (train/dev/test) used in Natural Questions-Open for the reading comprehension setting as well.
During the validation of our model and baseline models, we exclude samples whose answers lie in a list or a table from a Wikipedia article.

\section{Hyperparameters}\label{apdx:hyper}
We use the Adam optimizer~\citep{kingma2014adam} in all our experiments.
For training our phrase and question encoders with Eq.~\eqref{eqn:aggregate}, we use a learning rate of 3e-5 and the norm of the gradient is clipped at 1.
We use a batch size of $B=$84 and train each model for 4 epochs for all datasets, where the loss of pre-batch negatives is applied in the last two epochs.
% We found that setting $\lambda_1=1$, $\lambda_2=2$, and $\lambda_3=4$ works well in practice.
We use SQuAD to train our QG model\footnote{The quality of generated questions from a QG model trained on Natural Questions is worse due to the ambiguity of information-seeking questions.} and use spaCy\footnote{\url{https://spacy.io/}} for extracting named entities in each training passage, which are used to generate questions.
The number of generated questions is 327,302 and 1,126,354 for SQuAD and Natural Questions, respectively.
The number of preceding batches $C$ is set to 2.

For the query-side fine-tuning with Eq.~\eqref{eqn:qsft}, we use a learning rate of 3e-5 and the norm of the gradient is clipped at 1.
We use a batch size of 12 and train each model for 10 epochs for all datasets.
The top $k$ for the Eq.~\eqref{eqn:qsft} is set to 100.
While we use a single 24GB GPU (TITAN RTX) for training the phrase encoders with Eq.~\eqref{eqn:aggregate}, query-side fine-tuning is relatively cheap and uses a single 12GB GPU (TITAN Xp).
Using the development set, we select the best performing model (based on EM) for each dataset, which are then evaluated on each test set.
Since SpanBERT only supports cased models, we also truecase the questions~\citep{lita2003truecasing} that are originally provided in the lowercase (Natural Questions and WebQuestions).

% \section{Ablation Study}\label{apdx:ablation}
% In Table~\ref{tab:piqa-ablation}, we present ablation studies of \ours~in the reading comprehension setting (SQuAD).
% We observe that not sharing phrase and question encoders ({Share} = \xmark) and using the full output dimension ({Split} = \xmark) together improves the performance by 2\%.
% Unlike DenSPI, using a stronger pre-trained LM SpanBERT leads to another 1.3\% improvement in our model.
% Finally, augmenting the training set with generated questions ({QG} = \cmark) and performing distillation from query-dependent models ({Distill} = \cmark) further improves performance up to EM = 78.3.

% In Table~\ref{tab:qsft-ablation}, we show the effect of query-side fine-tuning on five open-domain QA datasets.
% We use three different phrase encoders, each of which is trained on a different training dataset $\traincorpus_\text{phrase}$.
% We find that query-side fine-tuning not only improves the performance on datasets that haven't been used as $\traincorpus_\text{phrase}$, but also helps improving the performance on datasets that have been used as $\traincorpus_\text{phrase}$ as well.

% \input{tables/semi_open_qa}
% \input{tables/piqa_ablation}
% \input{tables/qsft_ablation}

\section{Reducing Storage Footprint}\label{apdx:storage}
As shown in Table~\ref{tab:category}, we have reduced the storage footprint from 1,547GB~\citep{lee2020contextualized} to 320GB.
We detail how we can reduce the storage footprint in addition to the several techniques introduced by~\citet{seo2019real}.

First, following~\citet{seo2019real}, we apply a linear transformation on the passage token representations to obtain a set of filter logits, which can be used to filter many token representations from $\wordset$.
This filter layer is supervised by applying the binary cross entropy with the gold start/end positions (trained together with Eq.~\eqref{eqn:aggregate}).
We tune the threshold for the filter logits on the reading comprehension development set to the point where the performance does not drop significantly while maximally filtering tokens.
In the full Wikipedia setting, we filter about 75\% of tokens and store 770M token representations.

Second, in our architecture, we use a base model (SpanBERT-base) for a smaller dimension of token representations ($d=768$) and does not use any sparse representations including tf-idf or contextualized sparse representations~\citep{lee2020contextualized}.
We also use the scalar quantization for storing \ttt{float32} vectors as \ttt{int4} during indexing.

Lastly, since the inference in Eq.~\eqref{eqn:formula} is purely based on MIPS, we do not have to keep the original start and end vectors which takes about 500GB.
However, when we perform query-side fine-tuning, we need the original start and end vectors for reconstructing them to compute Eq.~\eqref{eqn:qsft} since (the on-disk version of) MIPS index only returns the top-$k$ scores and their indices, but not the vectors.

\end{document}